\documentclass{article}

\usepackage{microtype}
\usepackage{graphicx}
\usepackage{subfigure}
\usepackage{booktabs}
\usepackage{multirow}
\usepackage{amsmath}
\usepackage{amssymb}

\usepackage{hyperref}
\expandafter\def\expandafter\UrlBreaks\expandafter{\UrlBreaks\do\-}

\usepackage[accepted]{mlsys2025}
\usepackage{marvosym}

\usepackage{xcolor}
\usepackage{colortbl}
\newif\ifshowcomments
\showcommentstrue
\ifshowcomments
  \newcommand{\dayou}[1]{{\color{blue}[Dayou: #1]}}
  \newcommand{\jc}[1]{{\color{red}[Jianyi: #1]}}
\else
  \newcommand{\dayou}[1]{\ignorespaces}
  \newcommand{\jc}[1]{\ignorespaces}
\fi

\mlsystitlerunning{BitDecoding-FP4: Native FP4 KV Cache Decoding via Algorithm--System Co-Design}

\begin{document}

\twocolumn[
\mlsystitle{BitDecoding-FP4: Native FP4 KV Cache Decoding via Algorithm--System Co-Design}

\begin{mlsysauthorlist}
\textbf{DaYou Du}\textsuperscript{1}, \quad
\textbf{Woon Yee Ng}\textsuperscript{1}, \quad
\textbf{Guangming Sheng}\textsuperscript{2}, \quad
\textbf{Nick Lau}\textsuperscript{2}, \quad
\textbf{Jianyi Cheng}\textsuperscript{1}, \quad
\textbf{Luo Mai}\textsuperscript{1}
\\
\textsuperscript{1}University of Edinburgh \quad \textsuperscript{2}Tencent Hunyuan
\\
\Letter~\href{mailto:dayou.du@ed.ac.uk}{\texttt{dayou.du@ed.ac.uk}}
\end{mlsysauthorlist}

\mlsyskeywords{KV Cache Quantization, NVFP4, Long-Context LLM Inference}

\vskip 0.3in

\begin{abstract}
The key-value (KV) cache dominates memory capacity and bandwidth in long-context LLM serving.
While 4-bit microscaling formats (NVFP4, MXFP4) shrink the cache and run attention on FP4 Tensor
Cores, a fundamental gap remains between algorithmic quantization and system-aware acceleration:
existing FP4 KV systems use vanilla quantization that causes severe accuracy drops, whereas
accuracy-oriented 4-bit algorithms rely on vector codebooks whose lookup overhead degrades decoding
throughput. We present BitDecoding-FP4, an algorithm--system co-design for accurate and efficient
native FP4 KV decoding. We mitigate three error sources: Dual-Residual Query Quantization splits
queries into two FP4 tensors, cutting query logit error from 60\% to 1\%; Underflow-Aware
Probability Scaling prevents E4M3 scales from rounding long-context probabilities to zero; and
Outlier-Aware Key Quantization combines static smoothing, Hadamard rotation, and scale search to
suppress key outliers under token-wise scaling. At the system level, we remove the softmax
bottleneck via transposed softmax with fused probability quantization and stacked query MMA on
paged KV caches. Across dense and MoE models with GQA and MLA, BitDecoding-FP4 stays within 0.5
points of BF16 KV accuracy with NVFP4 (1 point with MXFP4), and tracks BF16 KV in RL training rollouts. In end-to-end serving on NVIDIA B300 with Qwen3-4B and Qwen3-30B-A3B, it delivers 2.4--3.0$\times$ higher peak serving throughput than BF16 KV.
The implementation is available at \url{https://github.com/OpenBitSys/BitDecoding-FP4}.
\end{abstract}
]


\section{Introduction}
\label{sec:introduction}

Modern large language models (LLMs) have rapidly expanded their context windows, with recent
frontier models supporting up to 1M tokens~\citep{team2024gemini15,yang2025qwen3}, enabling
complex applications such as long-document comprehension, repository-level coding, and test-time
reasoning scaling~\citep{wei2022chain,guo2025deepseekr1}. However, serving these models at long
contexts and large batch sizes is severely bottlenecked by the key-value (KV)
cache~\citep{kwon2023pagedattention,du2026bitdecoding}. Unlike static model weights, the KV cache
grows linearly with both sequence length and concurrency, quickly dominating GPU memory capacity and
bandwidth as it is repeatedly loaded during decoding~\citep{xia2026kitty,zhou2026oscar}.

Quantizing the KV cache to low precision effectively compresses its memory
footprint and reduces data movement~\citep{liu2024kivi,hooper2024kvquant}. Recently, 4-bit
microscaling formats such as MXFP4 and NVFP4~\citep{rouhani2023microscaling,nvidia2025ptx} have
improved low-bit numerical fidelity while enabling native 4-bit matrix multiplications directly
on Tensor Cores, delivering $2\times$ or higher throughput than FP8 on modern NVIDIA
Blackwell and Rubin GPUs~\citep{nvidia2025blackwellultra} and eliminating the runtime dequantization overhead of low-bit KV caches.
Recent works further show that these formats
achieve high accuracy and efficiency for weight and activation quantization in
training and inference~\citep{cook2025fouroversix,egiazarian2026bridging}, as well as for prefill and non-causal attention~\citep{zhang2025sageattention3,yu2026mxattention,zhang2026attn}.

\begin{figure}[!t]
  \centering
  \includegraphics[width=\columnwidth]{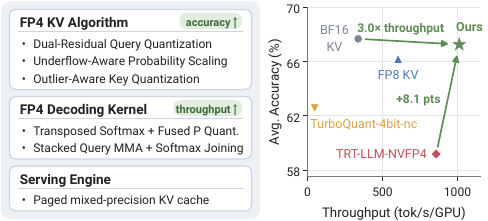}
  \caption{Overview of BitDecoding-FP4 (left), achieving higher accuracy and serving throughput
  when running Qwen3-4B (right). See Table~\ref{tab:main_accuracy} for the average accuracy and
  Figure~\ref{fig:e2e_128k_tp2} for the per-GPU serving throughput on B300 at 128K context.}
  \label{fig:intro}
\end{figure}

However, for autoregressive KV cache decoding, a fundamental gap remains between algorithmic quantization and system-aware acceleration, as shown in Figure~\ref{fig:intro} (right).
On the one hand, existing efficient NVFP4 KV cache systems such as TRT-LLM-NVFP4~\citep{nvidia2025trtllm} directly apply vanilla NVFP4 to the KV cache, causing severe accuracy drops in long-context inference; meanwhile, DeepInfer~\citep{deepseek2026v41flash, deepinfer} relies on post-training without considering the underlying quantization error.
On the other hand, accuracy-oriented KV quantization algorithms such as TurboQuant~\citep{zandieh2026turboquant} rely on vector codebooks whose online lookup and reconstruction overhead is inefficient on GPUs, severely degrading decoding throughput.

To further push decoding performance, executing attention \emph{natively} on FP4 Tensor Cores while preserving accuracy poses coupled algorithmic and system challenges.
First, at the algorithm level, whereas dequantization-based kernels keep queries $Q$ and softmax probabilities $P$ in 16 or 8 bits, native FP4 attention requires all four operands ($Q, K, P, V$) in narrow 4-bit microscaling blocks. During long-context decoding, query rounding noise is broadcast across thousands of tokens, small probabilities $P$ severely underflow under 8-bit E4M3 block scales, and token-wise Tensor Core block layouts clash with steep channel-wise outliers in $K$.
Second, at the system level, simply invoking FP4 Tensor Cores on paged KV caches~\citep{kwon2023pagedattention,du2026bitdecoding,sun2026janusquant} does not automatically yield decoding speedups: as matrix multiplication accelerates, the bottleneck shifts to CUDA-core softmax reductions, and accuracy-preserving quantization steps (e.g., per-tile FP32 rescaling or extra query GEMMs~\citep{zhang2025sageattention3}) stall the decoding pipeline if not co-designed with the kernel.

In this paper, we present BitDecoding-FP4, which addresses both challenges through an algorithm--system co-design for \emph{native} FP4 KV cache decoding, as illustrated in Figure~\ref{fig:intro} (left). 
At the algorithm level, we identify and mitigate the three primary error sources in native FP4 attention:
(i) \emph{Dual-Residual Query Quantization} represents the query as the sum of two FP4 tensors to suppress logit error from query quantization;
(ii) \emph{Underflow-Aware Probability Scaling} prevents E4M3 block scales from rounding small long-context softmax probabilities to zero via exponent rescaling and adaptive scale-maximum selection; and
(iii) \emph{Outlier-Aware Key Quantization} combines static key smoothing, Hadamard rotation, and block-scale search to tame key channel outliers under token-wise quantization.
At the kernel level, we remove the softmax bottleneck with a \emph{transposed softmax} that parallelizes reductions across keys and fuses probability quantization into warp-level butterfly shuffles, while \emph{stacked query MMA} folds the residual query into a single narrow FP4 MMA and integrates seamlessly with paged KV caches~\citep{kwon2023pagedattention} via a device-side control kernel to reduce metadata overhead.

This paper makes the following contributions:
\begin{itemize}
  \item \textbf{Accuracy-preserving FP4 KV quantization algorithm.} 
  We propose a microscaling FP4 KV quantization recipe that stays within 0.5 points of BF16 KV in average accuracy with NVFP4 (within 1 point with MXFP4) across challenging long-context and reasoning benchmarks, outperforming FP8 KV and prior 4-bit methods, and generalizes across dense and MoE models up to 770B with GQA and MLA attention. In RL post-training rollouts, it closely tracks BF16 KV accuracy, avoiding the degradation of standard NVFP4.
  \item \textbf{Fast native FP4 decoding kernels.} 
  Our native FP4 attention kernels keep the accuracy-recipe overhead within 4\% and achieve up to $2.99\times$ and $4.01\times$ speedups over BF16 KV at 128K context on B300 and RTX PRO 5000, outperforming TRT-LLM-NVFP4, FlashInfer-XQA-NVFP4, and BitDecoding.
  \item \textbf{High end-to-end serving efficiency.} 
  In end-to-end serving, BitDecoding-FP4 shrinks KV cache memory by up to $3.5\times$, cuts decoding latency by up to $1.81\times$, and delivers $2.4\text{--}3.0\times$ higher peak throughput than BF16 KV at 64K--128K contexts, matching or exceeding TRT-LLM-NVFP4.
\end{itemize}

\section{Background and Motivation}
\label{sec:background}


\paragraph{Microscaling FP4 formats.}
Microscaling formats~\citep{rouhani2023microscaling} partition an input tensor $x$ into contiguous blocks $x_b \in \mathbb{R}^B$ of size $B$, storing each block as $B$ low-bit quantized elements $\hat{x}_b$ together with a single shared scalar scale $s_b$. For 4-bit floating-point (FP4), each quantized element takes values from the E2M1 grid
$\mathcal{G}_{\mathrm{E2M1}} = \{0, \pm 0.5, \pm 1, \pm 1.5, \pm 2, \pm 3, \pm 4, \pm 6\}$.
Given a block $x_b$, the block quantizer $\phi(x_b) = (s_b, \hat{x}_b)$ computes 
\begin{equation}
  s_b = \frac{\max_i |x_{b,i}|}{6}, \qquad
  \hat{x}_b = \mathrm{round}_{\mathrm{E2M1}}\big(x_b / s_b\big),
  \label{eq:mx-quant}
\end{equation}
where $\max_i |x_{b,i}|$ is the maximum absolute value in $x_b$ and $\mathrm{round}_{\mathrm{E2M1}}(\cdot)$ rounds each element to its nearest value in $\mathcal{G}_{\mathrm{E2M1}}$, mapping the block's largest magnitude to $6$ (the maximum representable E2M1 value). Collecting $(s_b, \hat{x}_b)$ across all blocks yields the quantized tensor $\phi(x) = (s, \hat{x})$. Dequantization multiplies the
scales back, $\phi^{-1}(s, \hat{x}) = s \odot \hat{x}$, where $\odot$ broadcasts each block scale $s_b$ across the $B$ elements of $\hat{x}_b$, and we write $Q(x) = \phi^{-1}(\phi(x))$ for the quantize--dequantize round trip. NVFP4
uses block size $B = 16$ with an 8-bit E4M3 scale $s_b$, whereas MXFP4 uses $B = 32$ with a power-of-two E8M0 scale,
so $s_b$ is rounded to a power of two. Both formats natively carry fine-grained scales over 16 or 32 elements to reduce rounding error, and their E2M1 grid, which is densest
near zero, fits bell-shaped tensor distributions better than the evenly spaced INT4 grid.

\paragraph{Accuracy drops in FP4 KV cache decoding.}
With adaptive scale selection and rotation, microscaling FP4 now quantizes weights and activations
accurately~\citep{cook2025fouroversix,egiazarian2026bridging}, but the KV cache remains a gap.
TensorRT-LLM~\citep{nvidia2025trtllm} and DeepSeek-V4.1-Flash~\citep{deepseek2026v41flash} store
the KV cache in NVFP4 with the standard quantizer of Eq.~\ref{eq:mx-quant}. In our evaluation,
TRT-LLM-NVFP4 loses 8.5 average points against BF16 KV on Qwen3-4B (\mbox{Table~\ref{tab:main_accuracy}}). Native FP4 attention methods such as
\mbox{SageAttention3}~\citep{zhang2025sageattention3} and Attn-QAT~\citep{zhang2026attn} do not close this
gap either: they target prefill and diffusion workloads, which quantize BF16 queries, keys, and
values on the fly within each call, rather than decoding from a low-precision KV cache.

\begin{figure}[t]
  \centering
  \includegraphics{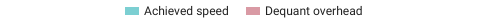}\\[2pt]
  \subfigure[Qwen3-4B]{\includegraphics{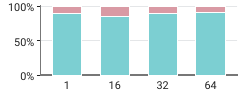}\label{fig:dequant-overhead-4b}}\hfill
  \subfigure[Qwen3-30B-A3B]{\includegraphics{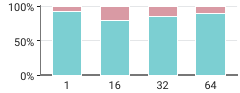}\label{fig:dequant-overhead-30b}}
  \caption{Dequantization overhead of FlashInfer-XQA~\citep{ye2025flashinfer} kernel with NVFP4 KV Cache.}
  \label{fig:dequant-overhead}
\end{figure}

\paragraph{Pushing performance limits via native FP4 execution.}
Most low-bit KV cache kernels dequantize the cache to BF16~\citep{lin2025qserve,ye2025flashinfer,du2026bitdecoding}
or FP8~\citep{nvidia2025trtllm} on CUDA cores before each attention matrix multiplication, leaving
Tensor Cores underutilized: for an NVFP4 cache, dequantization takes
7--21\% of FlashInfer-XQA's decode kernel time at 128K context (Figure~\ref{fig:dequant-overhead}).
NVIDIA Blackwell GPUs (e.g., B300, RTX PRO 5000, and RTX 5090) remove this step with
fifth-generation Tensor Cores that natively multiply block-scaled NVFP4 and MXFP4
operands~\citep{nvidia2025ptx} at $2\text{--}3\times$ the throughput of FP8, consuming the E2M1
elements and block scales directly as $\mathrm{FP4MM}(\hat{x}, s_x, \hat{y}, s_y)$.
For attention execution, given a query $q \in \mathbb{R}^d$ and cached $K, V \in \mathbb{R}^{N \times d}$ (with
token rows $k_i, v_i \in \mathbb{R}^d$), decoding computes attention scores (logits)
$S = \gamma\, q K^\top$ with per-token logit $S_i = \gamma\, q^\top k_i$ ($\gamma = \log_2 e / \sqrt{d}$),
probabilities $P = \mathrm{softmax}(S)$ via online softmax~\citep{milakov2018online,dao2022flashattention}, and output
$o = P V$; native FP4 attention~\citep{zhang2025sageattention3} quantizes all four operands into
microscaling FP4 to execute both products via $\mathrm{FP4MM}$.

In this work, we build upon the Tensor Core-aware KV layout and mixed-precision buffer of BitDecoding~\citep{du2026bitdecoding}, and focus on an accurate microscaling FP4 KV quantization algorithm (Section~\ref{sec:algorithm}) and native FP4 Tensor Core kernels tailored to Blackwell GPUs (Section~\ref{sec:kernel}).




\section{FP4 KV Quantization Algorithm}
\label{sec:algorithm}

\newcommand{\fpstep}[1]{{\color{blue!75!black}#1}}
\newlength{\alglabw}
\makeatletter
\newcommand{\alglinelabel}[1]{\protected@edef\@currentlabel{\arabic{ALC@line}}\label{#1}}
\makeatother


In this section, we present the BitDecoding-FP4 decoding algorithm in Algorithm~\ref{alg:overview}, where both attention matrix multiplications run natively on FP4 Tensor Cores and our algorithmic modifications over standard FlashAttention~\citep{dao2022flashattention} are \fpstep{highlighted}. In the following subsections, we analyze the quantization error of each operand under microscaling FP4 and detail our corresponding techniques.



\begin{algorithm}[t]
  \caption{Overview of the BitDecoding-FP4 algorithm.}
  \label{alg:overview}
  \small
  \newcommand{\algnote}[2]{\item[]\vspace{3pt}\leavevmode{\color{gray}\hypersetup{linkcolor=gray}$\triangleright$~#1 (\S\ref{#2})}\vspace{3pt}}
  \settowidth{\alglabw}{\textbf{Require:}}
  \noindent\begin{tabular}{@{}l@{\hspace{0.5em}}>{\raggedright\arraybackslash}p{\dimexpr\linewidth-\alglabw-0.5em\relax}@{}}
    \textbf{Input:}   & Query $q \in \mathbb{R}^{d}$, $K, V \in \mathbb{R}^{N \times d}$, block size $B_k$. \\
    \textbf{Require:} & Hadamard matrix $H$, key mean $\mu_k$, constants $c$, $\tau$, and
                        \mbox{$\gamma = \log_2 e / \sqrt{d}$}. \\
    \textbf{Output:}  & Attention output $o$.
  \end{tabular}
  \vspace{-0.5\baselineskip}
  \begin{algorithmic}[1]
    \algnote{Outlier-Aware Key Quantization}{sec:key-quant}
    \STATE \textbf{Preprocessing:} \fpstep{$K' = (K - \mu_k)H^\top$};
    \STATE $(s_K, \hat{K}) = \phi(K')$, $(s_V, \hat{V}) = \phi(V)$;
    \STATE \fpstep{$q' = Hq$}, $(s^{(1)}, \hat{q}^{(1)}) = \phi(q')$;
    \algnote{Dual-Residual Query Quantization}{sec:query-splitting}
    \STATE \fpstep{$(s^{(2)}, \hat{q}^{(2)}) = \phi(q' - s^{(1)} \odot \hat{q}^{(1)})$};
    \STATE Divide $\hat{K}$, $\hat{V}$ and their scales into $T = N / B_k$ blocks $\{\hat{K}_j, s_{K_j}\}$, $\{\hat{V}_j, s_{V_j}\}$;
    \FOR{$j$ in $[1, T]$}
      \STATE $S_j = \gamma \cdot \mathrm{FP4MM}(\hat{q}^{(1)}, s^{(1)}, \hat{K}_j, s_{K_j})$;
      \STATE \fpstep{$S_j \mathrel{+}= \gamma \cdot \mathrm{FP4MM}(\hat{q}^{(2)}, s^{(2)}, \hat{K}_j, s_{K_j})$};
      \STATE $m_j = \max(m_{j-1}, \max(S_j))$;
      \algnote{Underflow-Aware Probability Scaling}{sec:p-scaling}
      \STATE $P_j = \exp_2(S_j - m_j \fpstep{{} + \log_2 c})$;\alglinelabel{alg:line-exp}
      \STATE $\ell_j = 2^{\,m_{j-1} - m_j}\,\ell_{j-1} + \mathrm{sum}(P_j)$;
      \STATE \fpstep{$p^{\max} = \exp_2(\mathrm{blockmax}(S_j) - m_j)$};
      \STATE \fpstep{$\kappa_P = 2$ \textbf{if} $p^{\max} \le \tau / c$ \textbf{else} $6$};%
        \alglinelabel{alg:line-db}
      \STATE $s_P = \fpstep{c\, p^{\max} / \kappa_P}$,
        $\hat{P}_j = \mathrm{round}_{\mathrm{E2M1}}(P_j / s_P)$;
      \STATE $o_j = 2^{\,m_{j-1} - m_j}\,o_{j-1} + \mathrm{FP4MM}(\hat{P}_j, s_P, \hat{V}_j, s_{V_j})$;
    \ENDFOR
    \STATE \textbf{return} $o = o_T / \ell_T$
  \end{algorithmic}
\end{algorithm}

\subsection{Dual-Residual Query Quantization}
\label{sec:query-splitting}

\begin{figure}[t]
  \centering
  \includegraphics{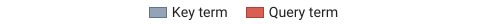}\\[2pt]
  \subfigure[NVFP4 keys]{\includegraphics{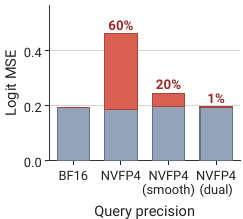}\label{fig:query-floor-nvfp4}}\hfill
  \subfigure[MXFP4 keys]{\includegraphics{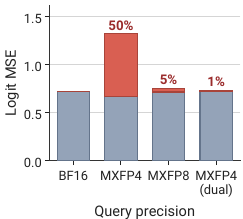}\label{fig:query-floor-mxfp4}}
  \caption{Mean squared logit error for each query format, split into key and query terms.
  Dual-Residual Query Quantization reduces the query quantization error to 1\% of the total.}
  \label{fig:query-floor}
\end{figure}

\paragraph{Logit error from query quantization.}

Native FP4 Attention requires the quantization of both query and keys. Writing $\tilde{x} = Q(x)$ for a quantized tensor and $\delta_x = \tilde{x} - x$ for its error, the resulting logit error $\tilde{S}_i - S_i$, up to the constant factor $\gamma$, decomposes into two terms: 

\begin{equation}
  \tilde{q}^\top \tilde{k}_i - q^\top k_i
  \;=\; \underbrace{\tilde{q}^\top \delta_{k_i} \vphantom{\delta_q^\top k_i}}_{\text{key error}\vphantom{qk}}
  \;+\; \underbrace{\delta_q^\top k_i \vphantom{\tilde{q}^\top \delta_{k_i}}}_{\text{query error}\vphantom{qk}} .
  \label{eq:logit-error}
\end{equation}

The attention output is highly sensitive to the logit error, as the softmax turns a small logit
perturbation into a shift of the attention distribution~\citep{hooper2024kvquant,li2025kvtuner}.
While the large quantization error brought by key outliers has been widely reported~\citep{liu2024kivi,zhou2026oscar},
we observe that query quantization also presents a challenge to attention accuracy.
Figure~\ref{fig:query-floor} confirms this by splitting the logit error into its key and query terms
for different query formats, showing that quantizing the query is not a second-order effect. With NVFP4 keys, the
NVFP4 query contributes 60\% of the logit error, more than the keys themselves.

\paragraph{Dual-residual quantization.}
To reduce this accuracy loss, we propose Dual-Residual Query Quantization, which represents the query as the
sum of two FP4 tensors. The first term is the ordinary quantized query; the second term quantizes
what the first term left behind:
\begin{equation}
  q \;\approx\;
  \underbrace{s^{(1)} \odot \hat{q}^{(1)}}_{q^{(1)} = Q(q)}
  \;+\;
  \underbrace{s^{(2)} \odot \hat{q}^{(2)}}_{q^{(2)} = Q(q - q^{(1)})} .
  \label{eq:split}
\end{equation}
where $s^{(1)}$ and $s^{(2)}$ are two separate sets of block scales (Eq.~\ref{eq:mx-quant}), computed
from the query $q$ and from the residual $q - q^{(1)}$, respectively. The query error in Eq.~\ref{eq:logit-error} shrinks from $\delta_q$ to the
residual's own rounding error $q - q^{(1)} - q^{(2)}$. Section~\ref{sec:kernel-query-split} explains
how the kernel computes the logits from the two terms.

\underline{Empirical results.}
We find that microscaling FP4 represents the residual query well: with NVFP4 keys, the dual-residual
query shrinks the query term from 60\% to only 1\% of the logit error (Figure~\ref{fig:query-floor}).
It also outperforms smoothing Q~\citep{zhang2025sageattention2,zhang2025sageattention3}, which still
leaves 20\% of the error to the query with a mean over the GQA heads and 42\% with a mean over earlier
tokens (Appendix~\ref{app:sageattention3}). With MXFP4 keys, the dual-residual query likewise leaves 1\%, whereas an
MXFP8 query, which current Tensor Cores natively accept against MXFP4 keys, still leaves 5\%.

\subsection{Underflow-Aware Probability Scaling}
\label{sec:p-scaling}

\begin{figure}[t]
  \centering
  \includegraphics{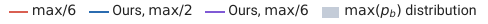}\\[2pt]
  \subfigure[E4M3 scale ladder]{\includegraphics{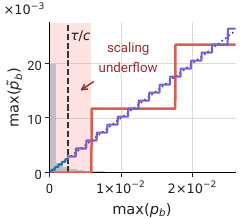}\label{fig:p-underflow-ladder}}\hfill
  \subfigure[Quantization error]{\includegraphics{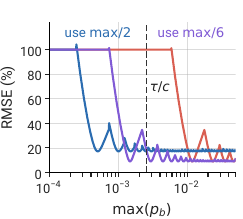}\label{fig:p-underflow-error}}
  \caption{(a) E4M3 scaling underflow on probability blocks in long-context decoding over the
  empirical block distribution. (b) Relative block quantization error (RMSE) across $\max(p_b)$.}
  \label{fig:p-underflow}
\end{figure}

\paragraph{Scaling underflow for $P$.}
We analyzed and observed that NVFP4 quantization of the softmax probabilities causes a large accuracy loss due to scaling underflow, especially in long-context decoding.
Since online softmax computes $p_i = 2^{\,S_i - m}$ from each token's logit $S_i$ and the running maximum logit $m$, almost every microscaling block along the sequence has a tiny maximum, $\max(p_b) \ll 1$. The scales of these blocks, $s_b = \max(p_b)/6$, when below $2^{-10}$, fall off the bottom of the E4M3 scaling range. On long reasoning traces, we observe that nearly 90\% of blocks have $\max(p_b) < 6 \cdot 2^{-10}$ and get $s_b = 0$ (Figure~\ref{fig:p-underflow}(a)), which means that those tokens are never considered because their blocks are always zero, especially hurting the retrieval nature of long contexts.

\paragraph{Underflow-aware scaling.}
To address this, we propose Underflow-Aware Probability Scaling, which lifts block scales away from the E4M3 floor with
two techniques. First, following SageAttention3~\citep{zhang2025sageattention3}, we rescale $p$ by a
constant factor $c$ before quantization, so that small blocks receive representable E4M3 scales.
Instead of re-multiplying every probability by an FP32 factor, we fold $c = 8$ into the exponent of
online softmax, $c\,p_i = 2^{\,S_i - m + \log_2 c}$, so the rescaling is free and cancels in the
softmax normalization. 
Second, we scale some blocks to the scale maximum $2$ instead of the largest E2M1 value $6$. In this
case, we can triple $s_b$ and further lift it away from the zero and subnormal floor of E4M3, while
also making the error distribution more even~\citep{cook2025fouroversix}. The scaling maximum
selection follows:
\begin{equation}
  s_b = \frac{c \max(p_b)}{\kappa_b}, \qquad
  \kappa_b =
  \begin{cases}
    2, & \max(p_b) \le \tau / c, \\
    6, & \text{otherwise}.
  \end{cases}
  \label{eq:p-scale}
\end{equation}

The threshold $\tau$ is calibrated offline on a small dataset to switch to
$\max/6$ once its subnormal E4M3 rounding spikes fall below the $\max/2$ error
(Figure~\ref{fig:p-underflow}(b)); calibration details are in Appendix~\ref{app:p-scale-calibration}. Each block is then quantized as
$\hat{p}_b = \mathrm{round}_{\mathrm{E2M1}}(c\,p_b / s_b)$. With both techniques, we maintain low
relative quantization error across the subnormal range and push the underflow floor down to
$\max(p_b) < 2^{-12}$, $24\times$ lower than under the standard rule
(Figure~\ref{fig:p-underflow}(a)). Section~\ref{sec:softmax} explains how this rule is fused into online softmax.

\begin{figure*}[t]
  \centering
  \subfigure[Roofline of decode attention]{%
    \includegraphics[height=1.42in]{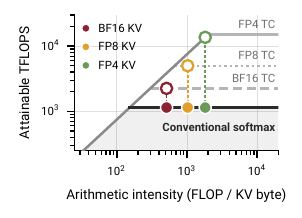}\label{fig:roofline}}\hfill
  \subfigure[Conventional softmax]{%
    \includegraphics[height=1.42in]{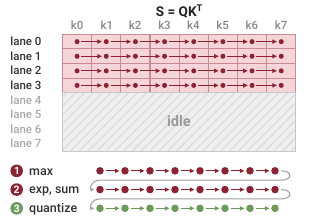}\label{fig:softmax-conventional}}\hfill
  \subfigure[Transposed softmax with fused P quantization]{%
    \includegraphics[height=1.42in]{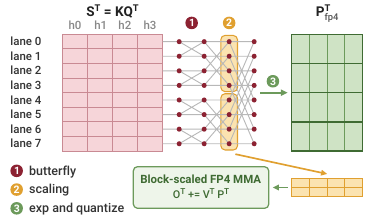}\label{fig:softmax-transposed}}
  \caption{(a) Roofline of decode attention on B300 (TC: Tensor Core): the conventional softmax caps every KV precision. (b) Conventional softmax: threads are
  assigned to head rows, so most sit idle. (c) Ours: threads are assigned to keys, so all threads do useful
  work, reductions run in parallel via butterfly shuffles, and the FP4 block scales come for free from
  the butterfly.}
  \label{fig:softmax}
\end{figure*}

\subsection{Outlier-Aware Key Quantization}
\label{sec:key-quant}

\paragraph{Channel outliers under token-wise key scaling.}
Key tensors exhibit persistent, large-magnitude outliers concentrated in a few fixed channels
across tokens~\citep{liu2024kivi,hooper2024kvquant}. Grouping keys channel-wise along the sequence
dimension can effectively isolate these outliers into their own scales. However,
native FP4 Tensor Cores compute each logit $S_i = \gamma\, q^\top k_i$ as a dot product across the $d$
channels of a single token $k_i$, and apply one shared scale to every 16-channel (NVFP4) or 32-channel
(MXFP4) slice of that dot product. Consequently, each key $k_i$ must be quantized \emph{token-wise}
along the channel dimension rather than channel-wise across tokens. Under token-wise scaling, a single
outlier channel dominates $\max |k_{i,b}|$ and inflates the block scale $s_b$, crushing the remaining
channels in the block and leading to severe quantization error.

\paragraph{Static key smoothing and Hadamard rotation.}
To suppress channel outliers under token-wise scaling, we first apply key smoothing, which removes
the static channel bias without extra computation due to the shift-invariant nature of
softmax~\citep{zhang2025sageattention2}. Unlike prefill attention kernels that compute the channel
mean $\mu_k$ dynamically over each sequence, we calibrate a static per-head mean $\mu_k$ offline
from eight prompts across the full context window and subtract it from each key before caching;
calibration details are in Appendix~\ref{app:key-smoothing-calibration}. Second, while smoothing
removes the static channel bias, dynamic token-dependent spikes remain. We further apply a
normalized Walsh--Hadamard rotation $H \in \mathbb{R}^{d \times d}$ ($H^\top H = I$) to both the
query and the smoothed key~\citep{zhou2026oscar}:
\begin{equation}
  q' = H q, \qquad k'_i = H(k_i - \mu_k),
  \label{eq:key-transform}
\end{equation}
which disperses residual channel spikes evenly across all $d$ coordinates while preserving the inner
product $q'^\top k'_i$.

\paragraph{Adaptive block-scale search.}
Finally, we select each block scale $s_b$ of the transformed key $k'_i$ from a two-candidate set
$\mathcal{S}_b$ to minimize the block reconstruction error:
\begin{equation}
  s_b = \arg\min_{s \in \mathcal{S}_b} \big\| s \, \mathrm{round}_{\mathrm{E2M1}}(x_b / s) - x_b \big\|_2^2.
  \label{eq:key-scale-search}
\end{equation}
For NVFP4, we follow Four Over Six~\citep{cook2025fouroversix} and search over the E4M3 scales for
scale maxima $6$ and $4$. For MXFP4, where scales are powers of two, we search between the standard OCP
scale $2^{e_{\mathrm{ocp}}}$~\citep{rouhani2023microscaling} and the next power of two
$2^{e_{\mathrm{ocp}}+1}$ (Appendix~\ref{app:mxfp4-scale-search}).

\section{System Design and Implementation}
\label{sec:kernel}

Pairing an FP4 KV cache with native FP4 Tensor Cores substantially reduces the memory footprint and
bandwidth traffic of decoding, but turning those savings into wall-clock speedup is non-trivial.
In Section~\ref{sec:softmax}, we identify softmax and on-the-fly probability quantization as the
primary execution bottleneck of native FP4 decoding, and propose a transposed softmax with fused
$P$ quantization to remove it. In Section~\ref{sec:kernel-query-split}, we show how Dual-Residual
Query Quantization stacks its second query term into the zero padding of the first, so both terms share one
narrow FP4 MMA per key tile. Finally, in Section~\ref{sec:kernel-impl}, we describe how BF16
attention-sink and recent tokens join the FP4 pages in the same decode launch, and how
BitDecoding-FP4 integrates into vLLM for online serving.

\subsection{Transposed Softmax with Fused P Quantization}
\label{sec:softmax}

\paragraph{Compute bottleneck from softmax.} 
In FlashAttention-style kernels~\citep{dao2022flashattention,zadouri2026flashattention4}, the attention score (logit) tile of height $M$ is partitioned across softmax threads along the row dimension. For a tile of $B_k$ keys and the $G$ query heads
that share a KV head, the conventional layout computes $S = Q K^\top \in \mathbb{R}^{G \times B_k}$, so
each thread is assigned to a query-head row (Figure~\ref{fig:softmax-conventional}). During decoding, $G$ is
usually much smaller than the MMA tile height $M$,
so most threads sit idle on padded rows, while the few active threads must scan across the $B_k$ keys of their rows in
three serial passes: finding the row maximum, computing $\exp_2$ and the row sum, and scanning the row again to find each block's maximum and
dynamically quantize $P$ to FP4. This per-thread serial work becomes the primary bottleneck of decode attention as
the KV cache precision drops: fewer bytes per token raise the arithmetic intensity
(Figure~\ref{fig:roofline}), but the softmax, rather than the Tensor Cores, sets the ceiling, leaving
the FP4 Tensor Cores largely idle.

\paragraph{Parallelizing softmax across keys.}
To eliminate the idle threads and break the per-row serial chain, we swap the two matrix-multiplication
operands and compute the transposed score tile $S^\top \in \mathbb{R}^{B_k \times G}$ and column-wise
probability tile $P^\top \in \mathbb{R}^{B_k \times G}$ across the $B_k$ keys:
\begin{equation}
  S^\top = \tilde{K} \tilde{Q}^\top, \quad
  P^\top = \mathrm{softmax}\big(S^\top\big), \quad
  O^\top = \tilde{V}^\top \tilde{P}^\top .
  \label{eq:transposed-attn}
\end{equation}
By choosing the key tile $B_k = M$, the keys fill all $M$ rows of the MMA tile regardless of the group size $G$:
every thread now holds the scores of valid keys in registers
(Figure~\ref{fig:softmax-transposed}). The softmax reductions over keys become parallel warp-level
butterfly reductions (\texttt{shfl.sync.bfly}) rather than a $B_k$-step loop per
thread, followed by a single cross-warp reduction in shared memory. Each thread's serial work thus shrinks
from $B_k$ steps to a few butterfly steps, with minimal communication overhead.

\paragraph{Fusing P quantization.}
The microscaling block scale comes directly out of the butterfly execution: each block of $P^\top$
($B = 16$ keys for NVFP4, $B = 32$ for MXFP4) covers consecutive keys
(Figure~\ref{fig:softmax-transposed}), so the row-maximum reduction already yields each block's
maximum logit $S^{\max}_b$ as an intermediate result, and since $\exp_2$ is monotonic,
$\max(p_b) = 2^{\,S^{\max}_b - m}$ requires no extra pass over $P$. Furthermore, the Underflow-Aware
Probability Scaling rule (Eq.~\ref{eq:p-scale}) fuses in with little overhead: choosing $\kappa_b$ is a
single predicate on $\max(p_b)$, and $c$ is folded into the $\exp_2$ argument
(lines~\ref{alg:line-exp} and~\ref{alg:line-db} of Algorithm~\ref{alg:overview}). Each thread finally
scales and rounds its own probability to E2M1 in registers, producing $\hat{P}^\top$ for the
block-scaled FP4 MMA.

\begin{figure}[t]
  \centering
  \includegraphics[width=\columnwidth]{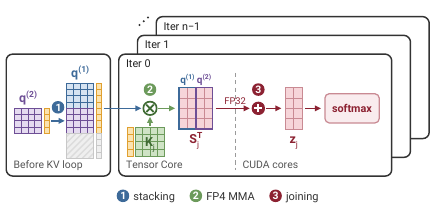}
  \caption{Stacked Query MMA with In-Softmax Joining. (1) Before the KV loop, $q^{(2)}$ is stacked
  into the zero padding of the $q^{(1)}$ tile. (2) Each KV block issues one FP4 MMA for both terms.
  (3) The softmax joins the two halves of $S_j^\top$ in FP32.}
  \label{fig:query-split}
\end{figure}

\subsection{Stacked Query MMA with In-Softmax Joining}
\label{sec:kernel-query-split}

\paragraph{Cost of a second query term.}
Dual-Residual Query Quantization (Section~\ref{sec:query-splitting}) represents the query as two FP4 terms
whose block scales differ widely, since $s^{(2)}$ is computed from the small residual $q - q^{(1)}$.
The two terms therefore cannot be merged into one FP4 operand: the kernel must form two block-scaled
products and add them in FP32,
\begin{equation}
  S^\top = \tilde{K}\,\tilde{Q}^{(1)\top} + \tilde{K}\,\tilde{Q}^{(2)\top} .
  \label{eq:split-scores}
\end{equation}
The direct implementation issues two MMAs per key tile that accumulate into the same FP32 scores.
In the transposed layout (Eq.~\ref{eq:transposed-attn}), the $G$ query heads form the column dimension of
each MMA; when $G$ is smaller than the MMA's minimum column width, the query operand is zero-padded, and the
second MMA largely repeats work on this padding.

\paragraph{Stacking in MMA and joining in softmax.}
Microscaling MMAs attach block scales to the rows of each operand, so different rows of one operand
can carry unrelated scales. We use this to place $\tilde{Q}^{(2)}$ in the zero padding of
$\tilde{Q}^{(1)}$ (Figure~\ref{fig:query-split}). Once TMA lands both query tiles, one load warp
copies the $G$ live rows of $\tilde{Q}^{(2)}$ and their block scales into rows $G$ to $2G-1$ of the
$\tilde{Q}^{(1)}$ tile and its scale tile, fences shared memory, and signals the MMA warp through an
mbarrier; since the query is fixed across the KV loop, this runs once per tile. Each key tile then
issues one FP4 MMA with $2G$ columns, which writes the two products of Eq.~\ref{eq:split-scores}
into adjacent columns of $S^\top$. Their FP32 addition moves into the softmax: each softmax thread,
which holds valid key scores (Section~\ref{sec:softmax}), loads $2G$ scores
and adds column $G + h$ to column $h$ in registers. Because this sum is formed before
masking and the row maximum, the online softmax operates on the full scores $S^\top$, so its running
maximum, rescaling, and normalization are unchanged.

\subsection{Mixed-Precision Online Serving Integration}
\label{sec:kernel-impl}

\begin{figure}[t]
  \centering
  \includegraphics[width=\columnwidth]{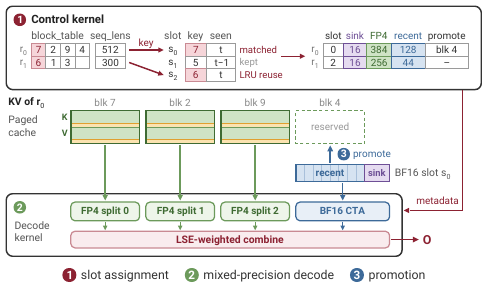}
  \caption{BitDecoding-FP4 in vLLM ($r_i$: batch row, $s_i$: BF16 slot). (1) The control kernel
  assigns each row a slot and splits its sequence. (2) The decode kernel reads the FP4 pages and the
  BF16 slot. (3) A full recent page is promoted into its FP4 block.}
  \label{fig:vllm-integration}
\end{figure}

\paragraph{Mixed-precision decoding.}
Following BitDecoding~\citep{du2026bitdecoding}, we keep the most recent tokens, which do not yet
fill a 128-token page, in BF16 and process them in the same Tensor Core pipeline as the FP4 pages,
while the first 16 tokens are kept as a permanent BF16 attention sink~\citep{xiao2024efficient}.
For long-context decoding, we split the FP4 pages of each request across CTAs as in
Flash-Decoding~\citep{dao2023flashdecoding}. Rather than attaching the BF16 sink and recent tokens
to one of these splits, making it the straggler, we merge them in an extra CTA alongside the FP4
splits to occupy otherwise idle SMs, and a log-sum-exp (LSE) weighted combine joins all
partial outputs.

\paragraph{Metadata management for paged KV caches.}
Serving engines such as vLLM~\citep{kwon2023pagedattention} allocate the KV cache in fixed-size
blocks on demand, and a per-request block table maps its pages to blocks. We store
each FP4 page and its block scales in one cache block (Figure~\ref{fig:vllm-integration}). The BF16
sink and recent tokens live outside the paged cache, in a buffer with one slot~$s_i$ per running
request. The engine is unaware of these slots and passes no request identity to the attention
kernel, so a lightweight control kernel manages them on the device at every step.
(1) The kernel identifies the request in each batch row~$r_i$ by the cache block holding its first
page, which stays fixed for the request's lifetime, and matches it against a slot table that
records the owner and last-seen step of each slot. In Figure~\ref{fig:vllm-integration}, row~$r_0$
finds its slot~$s_0$, while the new request in row~$r_1$ takes~$s_2$, the least recently seen slot
that no row claims; $s_1$ stays with its request, which is not batched in this step. From
the sequence lengths, the kernel then splits each row into its FP4 and recent tokens, 384 and 128
for~$r_0$, plus the 16 sink tokens, whose FP4 copy in the first page is masked.
(2) The decode kernel reads the FP4 pages through the block table and the BF16 tokens through
the slot of each row.
(3) Once the recent tokens fill a page, as for~$r_0$, the page is quantized into the cache block
already reserved for it, after the last layer has written the token that filled it.
Since this metadata never leaves the device, decoding requires no host synchronization and remains
compatible with CUDA graphs.


\section{Evaluations}
\label{sec:evaluation}

\newcommand{\oursrow}{\rowcolor{blue!6}}
\newcommand{\xspd}[1]{#1$\times$}

\begin{table*}[t]
  \centering
  \caption{Accuracy (\%) and end-to-end decoding latency of KV cache quantization methods;
  BitDecoding-NVFP4 stays within 0.5 average points of BF16 KV (BitDecoding-MXFP4 within 1 point)
  and outperforms FP8 and prior 4-bit methods in both average accuracy and decoding speed. LB-Pro denotes LongBench Pro, TPOT is the per-request
  time per output token (ms) at 256K context, and Speedup is relative to BF16 KV.}
  \label{tab:main_accuracy}
  \footnotesize
  \setlength{\tabcolsep}{3.6pt}
  \setlength{\aboverulesep}{0pt}
  \setlength{\belowrulesep}{0pt}
  \setlength{\extrarowheight}{0.6ex}
  \begin{tabular}{@{}ll ccc cc c c c|cc@{}}
    \toprule
    & & \multicolumn{3}{c}{Long context} & \multicolumn{2}{c}{Math} & Code & Knowl. & & \multicolumn{2}{c}{Latency @256K} \\
    \cmidrule(lr){3-5} \cmidrule(lr){6-7} \cmidrule(lr){8-8} \cmidrule(lr){9-9} \cmidrule(l){11-12}
    Model & Method & LB-Pro & MRCR & RULER & AIME24 & AIME25 & LCB v6 & GPQA & Avg. & TPOT & Speedup \\
    \midrule
    & BF16 KV           & 53.27 & 38.82 & 96.28 & 83.44 & 81.15 & 53.86 & 66.98 & 67.69 & 7.74  & \xspd{1.00} \\
    & FP8 KV            & 52.60 & 37.67 & 94.42 & 82.29 & 78.75 & 50.86 & 66.86 & 66.21 & 5.43  & \xspd{1.43} \\
    \cmidrule(l){2-12}
    & TurboQuant-4bit-nc  & 49.78 & 32.91 & 93.15 & 77.40 & 72.29 & 47.43 & 64.96 & 62.56 & 94.46 & \xspd{0.08} \\
    & SAW-INT4          & 49.92 & 31.57 & 91.19 & 79.69 & 75.63 & 52.86 & 64.02 & 63.55 & --$^*$ & -- \\
    & TRT-LLM-NVFP4      & 44.89 & 28.84 & 80.04 & 76.56 & 72.29 & 48.71 & 63.01 & 59.19 & 4.63  & \xspd{1.67} \\
    \oursrow \cellcolor{white} & BitDecoding-MXFP4 & 52.96 & 35.61 & 94.46 & 81.56 & 80.42 & 55.14 & 66.73 & 66.70 & 4.34 & \xspd{1.78} \\
    \oursrow \cellcolor{white} \multirow{-7}{*}{\shortstack[l]{Qwen3-4B-\\Thinking-\\2507}}
             & BitDecoding-NVFP4 & 52.34 & 36.68 & 95.68 & 82.50 & 80.62 & 56.14 & 66.92 & \textbf{67.27} & 4.28 & \textbf{\xspd{1.81}} \\
    \midrule
    & BF16 KV           & 60.83 & 43.41 & 98.39 & 90.62 & 83.54 & 64.00 & 71.28 & 73.15 & 6.66  & \xspd{1.00} \\
    & FP8 KV            & 58.45 & 40.60 & 98.03 & 91.56 & 82.81 & 60.86 & 70.58 & 71.84 & 5.15  & \xspd{1.29} \\
    \cmidrule(l){2-12}
    & TurboQuant-4bit-nc  & 59.56 & 42.22 & 98.02 & 88.65 & 83.85 & 56.57 & 70.64 & 71.36 & 126.55 & \xspd{0.05} \\
    & SAW-INT4          & 57.27 & 36.31 & 97.58 & 88.85 & 83.75 & 57.71 & 70.39 & 70.27 & --$^*$ & -- \\
    & TRT-LLM-NVFP4      & 50.72 & 28.41 & 89.42 & 88.54 & 81.77 & 61.00 & 70.90 & 67.25 & 4.60  & \xspd{1.45} \\
    \oursrow \cellcolor{white} & BitDecoding-MXFP4 & 58.53 & 42.71 & 98.33 & 91.04 & 84.58 & 62.29 & 70.71 & 72.60 & 4.59 & \xspd{1.45} \\
    \oursrow \cellcolor{white} \multirow{-7}{*}{\shortstack[l]{Qwen3-\\30B-A3B}}
             & BitDecoding-NVFP4 & 58.86 & 42.22 & 98.58 & 90.83 & 85.31 & 62.57 & 70.83 & \textbf{72.74} & 4.34 & \textbf{\xspd{1.53}} \\
    \bottomrule
    \multicolumn{12}{@{}l}{$^*$We omit the latency of SAW-INT4 because it runs on a different inference engine (SGLang) from all other methods (vLLM).} \\
  \end{tabular}
\end{table*}

Our evaluation highlights the following key results:
\begin{enumerate}
  \renewcommand{\labelenumi}{\arabic{enumi})}
  \setlength{\itemsep}{0pt}
  \setlength{\parsep}{0pt}
  \item \textbf{Accuracy:} within 0.5 average points of BF16 KV with NVFP4 (within 1 point with
  MXFP4), ahead of FP8 KV and prior 4-bit methods (Section~\ref{sec:eval-accuracy}).
  \item \textbf{Efficiency:} up to \xspd{4.01} faster kernels and \xspd{3.0} higher peak serving
  throughput than BF16 KV (Section~\ref{sec:eval-efficiency}).
  \item \textbf{Ablation:} the contribution of each algorithm and system technique
  (Section~\ref{sec:eval-ablation}).
\end{enumerate}

\subsection{Experiment Setup}

\paragraph{Models and Benchmarks.} We evaluate BitDecoding-FP4 on four models:
Qwen3-4B-Thinking-2507, Qwen3-30B-A3B-Thinking-2507~\cite{yang2025qwen3},
GLM-4.7-FP8~\cite{zeng2025glm45}, and Hy4-preview~\cite{tencent2026hy4preview}, covering dense and MoE models with
GQA and MLA attention. We evaluate long-context tasks with LongBench
Pro~\cite{chen2026longbenchpro}, MRCR~\cite{vodrahalli2024michelangelo,openai2025mrcr}, and
RULER-NIAH~\cite{hsieh2024ruler}, and long reasoning with AIME 2024/2025~\cite{maa2025aime},
LiveCodeBench v6~\cite{jain2025livecodebench}, and GPQA-Diamond~\cite{rein2024gpqa}.

\paragraph{Hardware.} Experiments run on NVIDIA B300 (SM103, 288GB) and RTX PRO 5000 (SM120, 72GB) GPUs. Accuracy and system-level evaluations use one B300 (TP=2 for 128K serving), and kernel benchmarks use both.

\paragraph{Implementation.} We implement BitDecoding-FP4 in CuTe-DSL~\cite{nvidia2025cutedsl}, built upon FlashAttention-4~\cite{zadouri2026flashattention4}, and integrate it into vLLM~\cite{kwon2023pagedattention} as a custom PagedAttention backend.

\paragraph{Baselines.} We compare against two higher-precision references and three 4-bit
KV cache methods. BF16 KV is the unquantized upper bound, and FP8 KV is vLLM's FP8 (E4M3)
KV cache. For 4-bit, TRT-LLM-NVFP4~\cite{nvidia2025trtllm} is the native NVFP4 KV cache in
vLLM, served by FlashInfer's TRT-LLM kernels; TurboQuant-4bit-nc~\cite{zandieh2026turboquant}
uses the official vLLM implementation, where for fairness we disable its mixed-precision
setup and quantize all layers; and SAW-INT4~\cite{jia2026saw} runs on SGLang.

\paragraph{RL Training.} To evaluate KV quantization during RL post-training, we follow DAPO~\cite{yu2025dapo} and run training with BF16 weights, using different KV cache methods for rollout generation and evaluating mean@4 accuracy on problems from DAPO, MATH-500, and AIME 2024.

\begin{table}[t]
  \centering
  \caption{Accuracy (\%) and KV cache memory footprint on large FP8-weight
  models with GQA and sparse MLA attention; BitDecoding-NVFP4 stays within 0.4 average points of BF16 KV while
  shrinking the KV cache by up to \xspd{3.5}.}
  \label{tab:large_models}
  \footnotesize
  \setlength{\tabcolsep}{4pt}
  \setlength{\aboverulesep}{0pt}
  \setlength{\belowrulesep}{0pt}
  \setlength{\extrarowheight}{0.6ex}
  \begin{tabular}{@{}l ccc|c@{}}
    \toprule
    Method & LB-Pro & MRCR & AIME25 & KV (GiB) \\
    \midrule
    \multicolumn{4}{@{}l|}{\textit{GLM-4.7 (355B, GQA), KV at 200K}} & \\
    BF16 KV           & 67.40 & 61.85 & 93.23 & 71.88 \\
    FP8 KV            & 65.87 & 61.27 & 92.40 & 35.94 \\
    TRT-LLM-NVFP4     & 62.28 & 56.29 & 91.88 & 20.21 \\
    \oursrow BitDecoding-NVFP4 & 66.74 & 60.83 & 93.75 & 20.27 \\
    \midrule
    \multicolumn{4}{@{}l|}{\textit{Hy4-preview (770B, MLA), KV at 1M}} & \\
    BF16 KV           & 60.29 & 67.96 & 98.65 & 90.46 \\
    FP8 KV            & 59.71 & 67.83 & 98.96 & 52.68 \\
    \oursrow BitDecoding-NVFP4 & 60.32 & 67.84 & 98.75 & 34.41 \\
    \bottomrule
  \end{tabular}
\end{table}

\subsection{Accuracy Results}
\label{sec:eval-accuracy}

\paragraph{Near-lossless accuracy across tasks.}
As shown in Table~\ref{tab:main_accuracy}, BitDecoding-NVFP4 stays within 0.5 average points of
BF16 KV on both Qwen3 models (67.27 vs.\ 67.69 and 72.74 vs.\ 73.15), and BitDecoding-MXFP4 stays within 1 point (66.70 and 72.60); notably,
both 4-bit variants surpass FP8 KV in average accuracy despite using half the bits. In contrast, existing 4-bit methods degrade substantially: native
TRT-LLM-NVFP4 loses up to 8.5 average points with sharp drops on long-context retrieval (RULER
and MRCR), while SAW-INT4 and TurboQuant-4bit-nc also fall behind.
On the 355B FP8-weight GLM-4.7 model (Table~\ref{tab:large_models}),
BitDecoding-NVFP4 stays within 0.4 average points of BF16 KV and shrinks the 200K-token KV cache by
\xspd{3.5} (71.88 to 20.27~GiB), whereas TRT-LLM-NVFP4 loses over 5 points on retrieval.

\begin{figure}[t]
  \centering
  \includegraphics{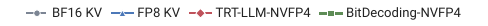}\\[2pt]
  \subfigure[Qwen3-4B]{\includegraphics{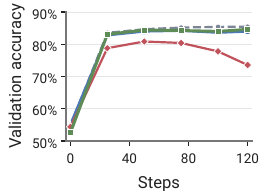}\label{fig:rl_qwen3_4b}}\hfill
  \subfigure[Qwen3-30B-A3B]{\includegraphics{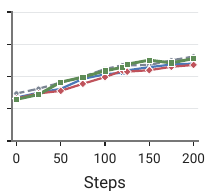}\label{fig:rl_qwen3_30b}}
  \caption{DAPO-Math RL validation accuracy with quantized rollout KV caches; BitDecoding-NVFP4 closely tracks BF16 KV.}
  \label{fig:rl_training_curves}
\end{figure}

\begin{figure*}[t]
  \centering
  \includegraphics{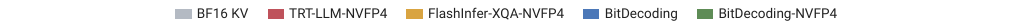}\\[2pt]
  \subfigure[B300 (SM103)]{\includegraphics{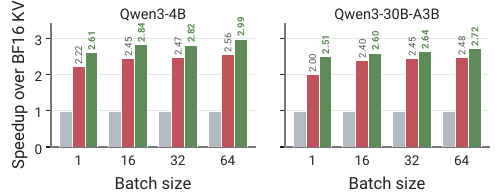}\label{fig:kernel_b300}}\hfill
  \subfigure[RTX PRO 5000 (SM120)]{\includegraphics{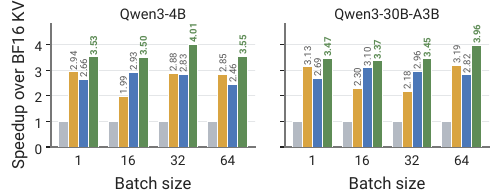}\label{fig:kernel_pro5000}}
  \caption{Decode attention kernel speedup over BF16 KV at 128K context across batch sizes
  (BitDecoding denotes the NVFP4 integration of~\cite{du2026bitdecoding});
  BitDecoding-NVFP4 consistently outperforms all baselines on both GPUs.}
  \label{fig:kernel_speedup}
\end{figure*}

\begin{figure*}[t]
  \centering
  \includegraphics{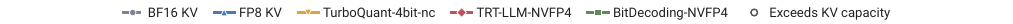}\\[2pt]
  \subfigure[64K in / 2K out, TP=1]{\includegraphics{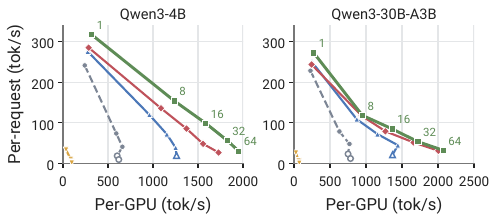}\label{fig:e2e_64k}}\hfill
  \subfigure[128K in / 2K out, TP=2]{\includegraphics{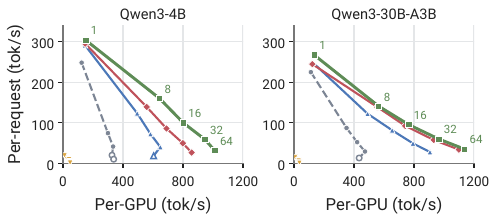}\label{fig:e2e_128k_tp2}}
  \caption{End-to-end serving throughput on Qwen3-4B and Qwen3-30B-A3B across 1, 8, 16, 32, and
  64 concurrent requests (up and to the right is better). When concurrent requests exceed KV cache
  capacity, excess requests are queued or preempted. BitDecoding-NVFP4 achieves the best
  latency--throughput trade-off across both workloads.}
  \label{fig:e2e_throughput}
\end{figure*}

\paragraph{Generalization to MLA.}
BitDecoding-NVFP4 also generalizes to MLA (implementation details in Appendix~\ref{app:nvfp4-mla}).
On the 770B FP8-weight Hy4-preview model (Table~\ref{tab:large_models}), it stays within 0.12 points of BF16 KV across all three tasks while cutting the 1M-token KV cache from 90.46 to 34.41~GiB (\xspd{2.6} smaller than BF16 KV and \xspd{1.5} smaller than FP8 KV).
Latency speedup is omitted because Hy4-preview uses sparse MLA with a fixed 2,048-token active window per decoding step, where attention compute is minimal and low-bit KV quantization primarily alleviates the 1M-token memory capacity bottleneck.

\paragraph{Stable RL with FP4 KV rollouts.}
\label{sec:rl_rollouts}
Recent models such as DeepSeek-V4.1-Flash~\citep{deepseek2026v41flash} adopt vanilla NVFP4 KV
caches during post-training without addressing FP4 error sources, even retaining FP8 for local sliding-window attention. As Figure~\ref{fig:rl_training_curves} shows, standard TRT-LLM-NVFP4
severely degrades RL training on long reasoning traces, whereas BitDecoding-NVFP4
closely tracks BF16 KV throughout training on both models.

\subsection{Efficiency Results}
\label{sec:eval-efficiency}

\paragraph{Faster decoding kernels.}
Figure~\ref{fig:kernel_speedup} compares decode kernel speedups over BF16 KV at 128K
context across batch sizes 1 to 64. On B300 (SM103), BitDecoding-NVFP4 achieves up to
\xspd{2.99} speedup on Qwen3-4B and \xspd{2.72} on Qwen3-30B-A3B, outperforming
TRT-LLM-NVFP4 across all batch sizes despite running our full accuracy
recipe. On RTX PRO 5000 (SM120), it reaches up to \xspd{4.01} and \xspd{3.96}
speedups, surpassing FlashInfer-XQA-NVFP4 and dequantization-based BitDecoding~\cite{du2026bitdecoding} by running both GEMMs natively on FP4 Tensor Cores.

\paragraph{End-to-end serving gains.}
At 256K context (Table~\ref{tab:main_accuracy}), BitDecoding-NVFP4 cuts per-request
decoding latency (TPOT) by \xspd{1.81} on Qwen3-4B and \xspd{1.53} on Qwen3-30B-A3B over BF16 KV,
outperforming FP8 KV and TRT-LLM-NVFP4, whereas TurboQuant-4bit-nc is over \xspd{10} slower due to vector-quantization overhead. Across 1 to 64 concurrent requests at 64K (TP=1) and 128K
(TP=2) contexts (Figure~\ref{fig:e2e_throughput}), BF16 and FP8 KV quickly exhaust KV capacity and suffer from preemption, whereas BitDecoding-NVFP4 fits all 64
requests in memory and delivers \xspd{2.4}--\xspd{3.0} higher peak per-GPU throughput
than BF16 KV, matching or exceeding TRT-LLM-NVFP4.

\begin{figure}[t]
  \centering
  \includegraphics{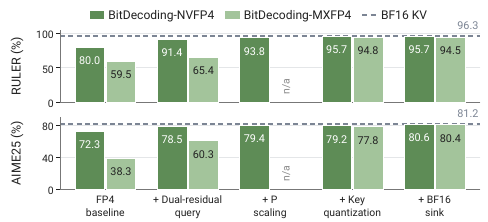}
  \caption{Accuracy ablation on Qwen3-4B (MXFP4 omits P scaling since E8M0 scales do not underflow).}
  \label{fig:ablation_accuracy}
\end{figure}


\subsection{Ablation Study}
\label{sec:eval-ablation}

\begin{figure}[!t]
  \centering
  \includegraphics{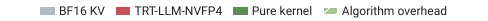}\\[2pt]
  \includegraphics{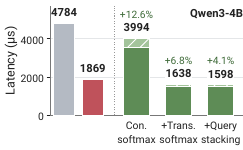}\hfill
  \includegraphics{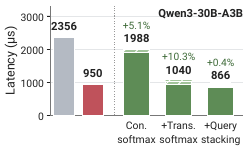}
  \caption{Decode kernel latency ablation at batch 64 and 128K context on B300.}
  \label{fig:kernel_breakdown}
\end{figure}

\paragraph{Improvement breakdown for accuracy.}
Figure~\ref{fig:ablation_accuracy} ablates our techniques (Section~\ref{sec:algorithm}) on Qwen3-4B. Dual-Residual Query Quantization and Outlier-Aware Key Quantization drive MXFP4 gains. For NVFP4, Dual-Residual Query Quantization recovers most of the gap, Underflow-Aware Probability Scaling restores RULER retrieval by preventing E4M3 underflow, and Outlier-Aware Key Quantization closes the rest (component ablations in Appendix~\ref{app:key-kl}).

\paragraph{Improvement breakdown for kernel speedup.}
Figure~\ref{fig:kernel_breakdown} breaks down decode kernel latency at batch 64 and 128K context. With conventional row softmax, softmax and $P$ quantization bottleneck execution (\xspd{2.1} slower than TRT-LLM-NVFP4 on \mbox{Qwen3-30B-A3B}). Transposed softmax (Section~\ref{sec:softmax}) nearly halves pure kernel latency (\xspd{1.91}), exposing the 10.3\% overhead of a separate residual-query MMA, and Stacked Query MMA (Section~\ref{sec:kernel-query-split}) folds both query terms into one narrow MMA, keeping recipe overhead within 4\% and outperforming TRT-LLM-NVFP4.

\begin{figure}[!t]
  \centering
  \includegraphics{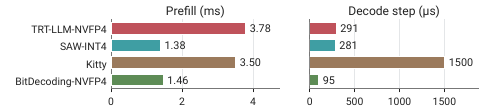}
  \caption{Quantization and metadata overhead of VLLM on Qwen3-4B. Both Kitty and BitDecoding-NVFP4 manage BF16 sink tokens.}
  \label{fig:metadata_overhead}
\end{figure}

\paragraph{Overhead of metadata management.}
At batch 1 and 4K context on Qwen3-4B (Figure~\ref{fig:metadata_overhead}), Kitty adds 1.5~ms per decode step for BF16 sink tokens, whereas our control kernel (Section~\ref{sec:kernel-impl}) keeps decode overhead to 95~$\mu$s (\xspd{15.8} below Kitty and \xspd{3} below TRT-LLM-NVFP4 and SAW-INT4), with prefill overhead (1.46~ms) matching SAW-INT4 and \xspd{2.6} below TRT-LLM-NVFP4.
\section{Related Works}

\paragraph{Microscaling FP4 for LLMs}

Microscaling FP4 formats~\citep{rouhani2023microscaling,nvidia2025ptx} have been widely adopted for
LLMs in post-training quantization~\citep{cook2025fouroversix,egiazarian2026bridging},
quantization-aware training~\citep{xin2026qad}, and FP4
training~\citep{castro2025quartet,panferov2026quartet2,nvidia2025nvfp4pretrain}, though these works
quantize only linear layers. For FP4 attention, Attn-QAT~\citep{zhang2026attn} relies on
quantization-aware training, and MXAttention~\citep{yu2026mxattention} quantizes unnormalized
exponentials to preserve softmax normalization under MXFP4. The most closely related post-training
work, SageAttention3~\citep{zhang2025sageattention3}, runs both attention matrix multiplications on
FP4 Tensor Cores via query/key smoothing~\citep{zhang2025sageattention,zhang2025sageattention2} and
two-level $P$ scaling (see algorithmic comparison in Appendix~\ref{app:sageattention3}). However,
these methods target prefill, non-causal, or training workloads (e.g., diffusion models) rather than
LLM decoding. 

\paragraph{Low-bit KV Caches for Decoding}

Recent algorithmic efforts explore 4-bit, 2-bit, and even 1-bit KV cache quantization~\citep{liu2024kivi,hooper2024kvquant,xia2026kitty,zandieh2025qjl}:
for instance, SpinQuant~\citep{liu2025spinquant} learns rotation matrices to suppress outliers, OScaR~\citep{su2026oscar} combines canalized rotation with omni-token scaling to address token norm imbalance, and TurboQuant~\citep{zandieh2026turboquant} pairs random rotation with near-optimal vector quantizers.
However, pushing to such extreme bit-widths still incurs non-negligible accuracy degradation on challenging long-context and reasoning tasks.
On the systems side, frameworks such as QServe~\citep{lin2025qserve}, SAW-INT4~\citep{jia2026saw}, Kitty~\citep{xia2026kitty}, and BitDecoding~\citep{du2026bitdecoding} accelerate low-bit KV cache decoding. While the original BitDecoding focuses on dequantization optimization, we co-design an accurate microscaling FP4 quantization algorithm with native FP4 Tensor Core kernels that avoid dequantization.
\section{Conclusion}
\label{sec:conclusion}

We presented BitDecoding-FP4, an algorithm--system co-design that bridges the gap between
accuracy-preserving quantization and hardware-native acceleration for microscaling FP4 KV cache
decoding. By mitigating the quantization errors of queries, attention probabilities, and keys
while eliminating softmax and quantization bottlenecks on FP4 Tensor Cores with paged KV caches,
BitDecoding-FP4 stays within 0.5 average points of BF16 KV with NVFP4 (1 point with MXFP4) across
models up to 770B, closely tracks BF16 KV in RL rollouts, and achieves up to $4.01\times$ kernel
speedup and $2.4\text{--}3.0\times$ higher peak serving throughput.

\section*{Acknowledgements}
We sincerely thank the University of Edinburgh for providing funding to rent the NVIDIA B300 GPU resources on Vast.ai that supported this research.

\bibliography{references}
\bibliographystyle{mlsys2025}

\clearpage
\appendix

\section{Teacher-Forced KL as a Fast Accuracy Proxy}
\label{app:forced-kl}

End-to-end benchmarks are slow to run and too noisy to separate configurations that are close
to BF16 KV. Once a quantized model samples a different token, its continuation departs from BF16
KV's, so most of the per-sample score difference comes from sampling rather than from quantization.
During development, we therefore measure each design choice by how
much it changes the model's next-token distribution along a fixed trajectory shared by all
configurations.

\paragraph{Measurement.}
The measurement takes two passes. First, BF16 KV generates a reference trajectory
$y = (y_1, \dots, y_L)$ for each prompt $x$ with the evaluation sampler, and records its next-token distribution
$p_t = p_{\text{BF16}}(\cdot \mid x, y_{<t})$ at every step. Second, each FP4 configuration
decodes the same prompts but is teacher-forced along $y$: at step $t$ it computes its own
distribution $\tilde{p}_t = p_{\text{FP4}}(\cdot \mid x, y_{<t})$ from its quantized KV cache, and
then emits $y_t$ instead of sampling. A vLLM logits processor forces the token, so the
configuration runs its unmodified serving path, including cache writes and the FP4 decode kernel.
The per-token divergence is
\begin{equation}
  D_t = \sum_{v \in \mathcal{V}_t} p_t(v) \log \frac{p_t(v)}{\tilde{p}_t(v)},
  \label{eq:forced-kl}
\end{equation}
where $\mathcal{V}_t$ holds BF16 KV's 128 most likely tokens at step $t$, which carry 99.996\% of
its probability mass on average. We average $D_t$ over the steps of each trajectory and then over
trajectories. Because all configurations see exactly the same prefixes, two of them can be compared
trajectory by trajectory: we report the paired $t$ statistic of their per-trajectory differences,
which cancels the large spread in difficulty across prompts.

\paragraph{Relation to end-to-end accuracy.}
Table~\ref{tab:kl-accuracy} places the KL of four configurations next to their MRCR accuracy. KL,
top-1 mismatch and accuracy rank the four configurations in the same order, and large reductions in
KL translate into large accuracy gains: BitDecoding-NVFP4 cuts the KL of TRT-LLM-NVFP4 by
10$\times$ and recovers 7.8 of its 10.0-point MRCR gap to BF16 KV.

\begin{table}[t]
  \centering
  \caption{Teacher-forced KL tracks end-to-end accuracy on Qwen3-4B: the lower a
  configuration's KL and top-1 mismatch on the MRCR trajectories, the higher its MRCR accuracy. KL
  is in units of $10^{-3}$ nats; top-1 mismatch is the fraction of steps where the FP4 argmax token
  differs from BF16 KV's, and the BF16 KV row reruns BF16 KV against its own reference.}
  \label{tab:kl-accuracy}
  \footnotesize
  \setlength{\tabcolsep}{5pt}
  \setlength{\aboverulesep}{0pt}
  \setlength{\belowrulesep}{0pt}
  \setlength{\extrarowheight}{0.6ex}
  \begin{tabular}{@{}l ccc@{}}
    \toprule
    Configuration & KL $\downarrow$ & Top-1 mis.\ (\%) $\downarrow$ & MRCR $\uparrow$ \\
    \midrule
    BF16 KV                   & 0.29  & 0.32 & 38.82 \\
    BitDecoding-NVFP4         & 3.83  & 1.00 & 36.68 \\
    BitDecoding-MXFP4         & 10.27 & 1.56 & 35.61 \\
    TRT-LLM-NVFP4             & 39.62 & 3.59 & 28.84 \\
    \bottomrule
  \end{tabular}
\end{table}

\section{KL Ablation of Key Smoothing Calibration}
\label{app:key-smoothing-calibration}

Key smoothing subtracts a static per-head key mean $\mu_k$ before caching (Section~\ref{sec:key-quant}).
We calibrate $\mu_k$ once per checkpoint and context window from only eight short prompts.

\paragraph{Calibration with eight prompts within one minute.}
The eight seed prompts are one-sentence requests on everyday topics, such as explaining hash tables
or photosynthesis, or writing a short story. The served model greedily continues each prompt for
256 tokens, so calibration generates only 2K tokens in total. We join the eight prompts and
continuations, tile them into one prompt that fills the whole context window (131,071 tokens for a
128K window), prefill it once, and accumulate the post-RoPE keys of every layer and KV head; their
mean is $\mu_k$. The text is the model's own output, so it matches what the model generates when
serving, and tiling lets a single prefill cover every position of the window, which matters
because RoPE makes the key mean depend on position. Generating only 2K tokens and running one
prefill keep calibration fast: it takes 13 seconds on one B300.

\begin{table}[t]
  \centering
  \caption{Key smoothing calibration on Qwen3-4B. Time is the seconds on one B300 GPU
  to produce the calibration text and measure $\mu_k$ with the serving engine. MRCR and AIME24 are
  the teacher-forced KL from BF16 KV ($\times 10^{-3}$, Appendix~\ref{app:forced-kl}), with the
  change relative to our calibration in parentheses.}
  \label{tab:key-smoothing-calibration}
  \footnotesize
  \setlength{\tabcolsep}{3pt}
  \setlength{\aboverulesep}{0pt}
  \setlength{\belowrulesep}{0pt}
  \setlength{\extrarowheight}{0.6ex}
  \begin{tabular}{@{}l rr cc@{}}
    \toprule
    Calibration & Tokens & Time & MRCR & AIME24 \\
    \midrule
    No smoothing   & --   & --  & 11.79 (+215\%) & 8.11 (+76\%) \\
    \midrule
    8 seeds (ours) & 133K & 13  & 3.74 & 4.62 \\
    8 other seeds  & 131K & 12  & 3.73 ($-$0.3\%) & 4.70 (+1.7\%) \\
    10 retrieval   & 1.3M & 43  & 4.52 (+21\%) & 5.27 (+14\%) \\
    40 mixed       & 5.2M & 173 & 4.35 (+16\%) & 4.88 (+5.5\%) \\
    \bottomrule
  \end{tabular}
\end{table}

\paragraph{Comparison with other calibration sets.}
Table~\ref{tab:key-smoothing-calibration} compares our $\mu_k$ with means calibrated on other text,
measured with the teacher-forced KL of Appendix~\ref{app:forced-kl} while the rest of the recipe
stays fixed. More calibration text does not help: ten retrieval prompts (1.3M tokens) and forty
prompts mixing synthetic retrieval, code, math and generated text (5.2M tokens) raise KL by
14--21\% and 5.5--16\%. Eight different seed prompts give the same KL on MRCR and 1.7\% higher on
AIME. 

\section{Probability Scale Threshold Calibration}
\label{app:p-scale-calibration}

We calibrate $\tau$ offline: replaying online softmax on the BF16 logits of decode queries, we select
the threshold that minimizes the total squared quantization error of $c\,p$ over all probability
blocks. Candidates lie on the E4M3 grid of $c \max(p_b)$, and the production $\tau = 21/1024$ is
code 10, between the E4M3 values $20/1024$ (code 10) and $22/1024$ (code 11).

\begin{table}[t]
  \centering
  \caption{Threshold code selected by each calibration source on Qwen3-4B-Thinking-2507; code 10 is
  the production $\tau$. The last four columns give the percentage of 1,000 random subsets of $N$
  samples that select code 10.}
  \label{tab:p-scale-calibration}
  \footnotesize
  \setlength{\tabcolsep}{3.5pt}
  \setlength{\aboverulesep}{0pt}
  \setlength{\belowrulesep}{0pt}
  \setlength{\extrarowheight}{0.6ex}
  \begin{tabular}{@{}l c rrrr@{}}
    \toprule
    & & \multicolumn{4}{c}{Subsets at code 10 (\%)} \\
    \cmidrule(l){3-6}
    Calibration source & Code & $N{=}1$ & 8 & 16 & 32 \\
    \midrule
    UUID key retrieval     & 10 & 100  & 100  & 100  & 100 \\
    Marker retrieval       & 10 & 100  & 100  & 100  & 100 \\
    Code and math filler   & 10 & 100  & 100  & 100  & 100 \\
    Key smoothing (ours)   & 10 & 100  & 100  & 100  & 100 \\
    Reasoning traces       & 10 & 97.9 & 84.8 & 70.2 & 100 \\
    \midrule
    All five sources       & 10 & 99.7 & 97.1 & 97.5 & 100 \\
    \bottomrule
  \end{tabular}
\end{table}

\paragraph{The threshold is consistent across calibration data.}
On Qwen3-4B-Thinking-2507, we build five calibration sources of 50 samples each, with contexts of 8K
to 128K tokens: three synthetic long-context families, the key smoothing text of
Appendix~\ref{app:key-smoothing-calibration}, and the model's own reasoning traces. Every source
selects code 10, and so does their union (Table~\ref{tab:p-scale-calibration}). Random subsets select
code 10 down to a single sample on every source except the reasoning traces, which need 32 traces for
every subset to select code 10.

\paragraph{One-time calibration with key smoothing.}
We therefore select $\tau$ on the key smoothing text, in the same one-time calibration as $\mu_k$
(Appendix~\ref{app:key-smoothing-calibration}): prefill keys of the text give $\mu_k$, and decode
queries that continue it give $\tau$. The two do not interact, since subtracting $\mu_k$ shifts all
logits of a query by the same $q^\top \mu_k$ and leaves $p$ unchanged. This text selects code 10 on
Qwen3-4B-Thinking-2507, and also on Qwen3-8B and Qwen3-30B-A3B-Thinking-2507 up to 32K tokens.

\section{Comparison with SageAttention3}
\label{app:sageattention3}

In this section, we compare smoothing Q~\citep{zhang2025sageattention2} and two-level scaling of $P$
from SageAttention3~\citep{zhang2025sageattention3} against our Dual-Residual Query Quantization and
Underflow-Aware Probability Scaling.

\begin{table}[t]
  \centering
  \caption{Mean squared logit error ($\times 10^{-2}$) of each query format against NVFP4 keys on
  Qwen3-4B. Dual-residual query quantization achieves the lowest error.}
  \label{tab:smooth-q}
  \footnotesize
  \setlength{\tabcolsep}{4pt}
  \setlength{\aboverulesep}{0pt}
  \setlength{\belowrulesep}{0pt}
  \setlength{\extrarowheight}{0.6ex}
  \begin{tabular}{@{}l ccc@{}}
    \toprule
    Query format & Total $\downarrow$ & Query term $\downarrow$ & Share \\
    \midrule
    BF16                             & 19.32 & 0.00  & --     \\
    NVFP4                            & 46.17 & 27.50 & 60\%   \\
    \quad + smoothing Q (token mean) & 33.03 & 13.95 & 42\%   \\
    \quad + smoothing Q (GQA mean)   & 24.57 & 4.89  & 20\%   \\
    \quad + dual residual (ours)     & 19.55 & 0.23  & 1.2\%  \\
    \bottomrule
  \end{tabular}
\end{table}

\paragraph{Comparison with smoothing Q.}
Smoothing Q subtracts a mean $\bar{q}$ from the query, quantizes the smaller remainder, and adds
$\bar{q}^\top \tilde{k}_i$ back to every logit in high precision using the dequantized keys. Decode has
a single query token per head, so we evaluate two choices of $\bar{q}$: the per-head mean of earlier
query tokens (token mean), and the mean over the $G$ query heads that share a KV head (GQA mean).

\underline{Higher accuracy.}
As shown in Table~\ref{tab:smooth-q}, Dual-Residual Query Quantization achieves the lowest logit error,
reducing the query term to just 1.2\% of the total, compared to 42\% for smoothing Q with the token mean
and 20\% with the stronger GQA mean (the variant shown in Figure~\ref{fig:query-floor}). Subtracting a
mean shrinks only the component shared across tokens or heads and still rounds the remainder to FP4 once, whereas
our second FP4 term directly quantizes that rounding error itself.

\underline{Higher efficiency.}
Smoothing Q requires extra CUDA-core work and instructions to compute the high-precision correction
$\bar{q}^\top \tilde{k}_i$ for each key, or extra memory overhead to store precomputed projections per
key. In contrast, our second query term stays on the Tensor Cores: stacked into the zero padding of the
first (Section~\ref{sec:kernel-query-split}), it only widens the query MMA from $G$ to $2G$ columns,
often within padding the MMA computes anyway, and is far more efficient.

\begin{table}[t]
  \centering
  \caption{Comparison of probability scaling rules on Qwen3-4B. Underflow-aware scaling matches the
  accuracy of two-level scaling without per-tile rescaling overhead.}
  \label{tab:two-level}
  \footnotesize
  \setlength{\tabcolsep}{4pt}
  \setlength{\aboverulesep}{0pt}
  \setlength{\belowrulesep}{0pt}
  \setlength{\extrarowheight}{0.6ex}
  \begin{tabular}{@{}l cc ccc@{}}
    \toprule
    $P$ scaling & KL $\downarrow$ & Top-1 $\uparrow$ & MRCR & RULER & LB-Pro \\
    \midrule
    BF16 KV                          & --   & --    & 36.08 & 90.25 & 55.79 \\
    Standard $\max/6$                & 9.31 & 98.50 & 35.53 & 82.00 & 54.99 \\
    Two-level                        & 6.06 & 98.75 & 35.51 & 88.00 & 55.12 \\
    Ours (Eq.~\ref{eq:p-scale})      & 6.47 & 98.73 & 35.51 & 88.75 & 55.02 \\
    \bottomrule
  \end{tabular}
\end{table}

\paragraph{Comparison with two-level scaling.}
SageAttention3 quantizes $P$ with two levels of scales: an FP32 scale per query row and key tile first
maps the tile maximum of $p$ to $448 \times 6$, and E4M3 block scales then quantize the result with the
standard $\max/6$ rule.

\underline{Similar accuracy.}
As shown in Table~\ref{tab:two-level}, our Underflow-Aware Probability Scaling achieves comparable KL
divergence, top-1 agreement, and downstream benchmark accuracy to two-level scaling, while both
substantially outperform the standard $\max/6$ rule.

\underline{Higher efficiency.}
Our constant $c = 8$ folds directly into the $\exp_2$ argument (line~\ref{alg:line-exp} of
Algorithm~\ref{alg:overview}) and cancels in the softmax normalization, adding no extra instructions or
output correction. In contrast, two-level scaling changes its FP32 scale at every tile, requiring exact
per-tile maximum reductions and per-tile output rescaling on CUDA cores. In our decode implementation of
two-level scaling, keeping the FP32 running output in registers for per-tile rescaling also causes
severe register spilling, making the kernel 3.0--26.7$\times$ slower than ours across 8K--128K contexts
and batches 1--16, and a serving decode step 5.0$\times$ slower (89.7 vs.\ 17.8\,ms).

\section{MXFP4 Adjacent-Exponent Block-Scale Search}
\label{app:mxfp4-scale-search}

In MXFP4~\citep{rouhani2023microscaling}, each block $x_b$ of 32 elements shares one unsigned 8-bit
power-of-two E8M0 scale $s_b = 2^e$ ($e \in [-127, 127]$). Because the largest representable E2M1
magnitude is $6 = 1.5 \times 2^2$, the standard OCP exponent rule sets
\begin{equation}
  e_{\mathrm{ocp}} = \lfloor \log_2(\max |x_b|) \rfloor - 2.
  \label{eq:mxfp4-ocp}
\end{equation}
Since $\max |x_b| \in [2^{\lfloor \log_2(\max |x_b|) \rfloor},\, 2^{\lfloor \log_2(\max |x_b|) \rfloor + 1})$,
dividing by $2^{e_{\mathrm{ocp}}}$ maps $\max |x_b|$ into $[4, 8)$. Whenever $\max |x_b| > 6 \cdot 2^{e_{\mathrm{ocp}}}$,
the standard OCP scale clips the block maximum down to $6 \cdot 2^{e_{\mathrm{ocp}}}$. Conversely,
selecting the next power of two, $e_{\mathrm{ocp}} + 1$, maps $\max |x_b|$ into $[2, 4)$ and avoids
clipping, at the cost of doubling the quantization step size for the rest of the block.

To balance clipping error on the largest elements against rounding error on the interior elements, we
evaluate both adjacent exponents for each block,
\begin{equation}
  \mathcal{S}_b = \big\{2^{e_{\mathrm{ocp}}},\, 2^{e_{\mathrm{ocp}} + 1}\big\},
  \label{eq:mxfp4-candidates}
\end{equation}
and select the scale that minimizes the squared reconstruction error in Eq.~\ref{eq:key-scale-search}
(breaking ties in favor of $2^{e_{\mathrm{ocp}}}$). Because the round-up exponent
$\lceil \log_2(\max |x_b| / 6) \rceil$ is always either $e_{\mathrm{ocp}}$ or $e_{\mathrm{ocp}} + 1$,
this two-candidate search is guaranteed to achieve no higher block MSE than either fixed rule.

\section{KL Ablation of Outlier-Aware Key Quantization}
\label{app:key-kl}

\begin{table}[t]
  \centering
  \caption{Leave-one-out KL ablation of Outlier-Aware Key Quantization on Qwen3-4B-Thinking-2507.
  Each row removes one part from the full recipe; values are the teacher-forced KL from BF16 KV
  ($\times 10^{-3}$, Appendix~\ref{app:forced-kl}). Since KL tracks end-to-end accuracy
  (Table~\ref{tab:kl-accuracy}), every part contributes to accuracy.}
  \label{tab:key-kl}
  \footnotesize
  \setlength{\tabcolsep}{5pt}
  \setlength{\aboverulesep}{0pt}
  \setlength{\belowrulesep}{0pt}
  \setlength{\extrarowheight}{0.6ex}
  \begin{tabular}{@{}l cc@{}}
    \toprule
    Configuration & MRCR & AIME24 \\
    \midrule
    NVFP4 full recipe             & 3.83 & 4.70 \\
    \quad $-$ key smoothing       & 11.79 (+208\%) & 8.11 (+72\%) \\
    \quad $-$ Hadamard rotation   & 6.68 (+74\%) & 6.84 (+45\%) \\
    \quad $-$ block-scale search  & 4.70 (+23\%) & 5.58 (+19\%) \\
    \midrule
    MXFP4 full recipe             & 10.27 & 11.75 \\
    \quad $-$ key smoothing       & 50.73 (+394\%) & 37.76 (+221\%) \\
    \quad $-$ Hadamard rotation   & 30.23 (+194\%) & 35.93 (+206\%) \\
    \quad $-$ block-scale search  & 12.97 (+26\%) & 17.19 (+46\%) \\
    \midrule
    BF16 KV (rerun)               & 0.29 & 0.45 \\
    \bottomrule
  \end{tabular}
\end{table}

Once the recipe is close to BF16 KV accuracy, end-to-end benchmarks cannot resolve the individual parts
of Outlier-Aware Key Quantization (Figure~\ref{fig:ablation_accuracy}). We therefore compare them
with the teacher-forced KL of Appendix~\ref{app:forced-kl}, on MRCR and AIME 2024 trajectories. Each row of
Table~\ref{tab:key-kl} changes one setting: it subtracts no key mean, applies no rotation to Q and K,
or keeps the single standard scale for K (amax/6 for NVFP4, the OCP exponent for MXFP4).

Every part matters in both formats. Removing key smoothing has the largest effect for both formats
on both benchmarks, tripling MRCR KL for NVFP4 and raising it almost fivefold for MXFP4. Removing the Hadamard
rotation raises KL by 45--74\% for NVFP4 and roughly triples it for MXFP4, and dropping block-scale
search adds 19--46\%. All twelve increases are significant: the paired $t$ is at least 10.2 over the
400 MRCR conversations and at least 14.6 over the 30 AIME problem means, and every AIME problem is
worse in every removal. For scale, the full NVFP4 recipe sits 13$\times$ above the BF16 KV floor on MRCR
and 10$\times$ on AIME 2024.

\section{FP4 Multi-Head Latent Attention}
\label{app:nvfp4-mla}

\begin{algorithm}[t]
  \caption{FP4 MLA decoding with a shared latent cache.}
  \label{alg:nvfp4-mla}
  \small
  \settowidth{\alglabw}{\textbf{Require:}}
  \noindent\begin{tabular}{@{}l@{\hspace{0.5em}}>{\raggedright\arraybackslash}p{\dimexpr\linewidth-\alglabw-0.5em\relax}@{}}
    \textbf{Input:}   & Content query $q \in \mathbb{R}^{d_k}$, positional query $r \in \mathbb{R}^{d_r}$,
                        latent cache $C \in \mathbb{R}^{N \times d_c}$, positional cache $R \in \mathbb{R}^{N \times d_r}$,
                        block size $B_k$. \\
    \textbf{Require:} & Projections $W_K, W_V$, Hadamard matrix $H$, latent mean $\mu_c$, and
                        $\gamma = \log_2 e / \sqrt{d_k+d_r}$. \\
    \textbf{Output:}  & Attention output $o \in \mathbb{R}^{d_v}$.
  \end{tabular}
  \vspace{-0.5\baselineskip}
  \begin{algorithmic}[1]
    \STATE \textbf{Preprocessing:} \fpstep{$C = (C - \mu_c)H^\top$}, $(s_C, \hat{C}) = \phi(C)$;
    \STATE $q_c = W_K^\top q$, \fpstep{$q_c = Hq_c$}, $(s^{(1)}, \hat{q}^{(1)}) = \phi(q_c)$;
    \STATE \fpstep{$(s^{(2)}, \hat{q}^{(2)}) = \phi(q_c - s^{(1)} \odot \hat{q}^{(1)})$};
    \STATE Divide $\hat{C}$, its scales, and $R$ into $T = N / B_k$ blocks $\{\hat{C}_j, s_{C_j}\}$ and $\{R_j\}$;
    \FOR{$j$ in $[1, T]$}
      \STATE $S_j = \gamma \cdot \mathrm{FP4MM}(\hat{q}^{(1)}, s^{(1)}, \hat{C}_j, s_{C_j})$;
      \STATE \fpstep{$S_j \mathrel{+}= \gamma \cdot \mathrm{FP4MM}(\hat{q}^{(2)}, s^{(2)}, \hat{C}_j, s_{C_j})$};
      \STATE $S_j \mathrel{+}= \gamma \cdot \mathrm{BF16MM}(r, R_j)$;
      \STATE $m_j = \max(m_{j-1}, \max(S_j))$;
      \STATE $P_j = \exp_2(S_j - m_j)$;
      \STATE $\ell_j = 2^{\,m_{j-1} - m_j}\,\ell_{j-1} + \mathrm{sum}(P_j)$;
      \STATE $\tilde{C}_j = \phi^{-1}(s_{C_j}, \hat{C}_j)$;
      \STATE $u_j = 2^{\,m_{j-1} - m_j}\,u_{j-1} + \mathrm{FP16MM}(P_j, \tilde{C}_j)$;
    \ENDFOR
    \STATE \fpstep{$\bar{u} = H^\top(u_T / \ell_T) + \mu_c$};
    \STATE \textbf{return} $o = W_V \bar{u}$
  \end{algorithmic}
\end{algorithm}

Algorithm~\ref{alg:nvfp4-mla} extends our FP4 decoding method to Multi-Head Latent Attention (MLA).
We present the formulation for a single query token and a single attention head, with the latent
cache shared across heads. Let $C \in \mathbb{R}^{N \times d_c}$ and $R \in \mathbb{R}^{N \times d_r}$
denote the cached latent and decoupled positional (RoPE) vectors over $N$ tokens. In MLA, the
per-head content keys and values are given by $K = C W_K^\top$ and $V = C W_V^\top$, where
$W_K \in \mathbb{R}^{d_k \times d_c}$ and $W_V \in \mathbb{R}^{d_v \times d_c}$ are the per-head
up-projection matrices, and the attention scale is $\gamma = \log_2 e / \sqrt{d_k + d_r}$.
During decoding, $W_K$ is absorbed into the content query as $q_c = W_K^\top q$, allowing content
scores to be computed directly against $C$, while $W_V$ is applied once to the aggregated latent
output rather than to every cached token.

Because the latent cache $C$ dominates the KV footprint, we apply Outlier-Aware Key Quantization
(Section~\ref{sec:key-quant}) directly to $C$ with latent mean $\mu_c \in \mathbb{R}^{d_c}$ and
Hadamard matrix $H \in \mathbb{R}^{d_c \times d_c}$, and apply Dual-Residual Query Quantization
(Section~\ref{sec:query-splitting}) to the rotated absorbed query $Hq_c$. The smaller positional
query $r \in \mathbb{R}^{d_r}$ and positional cache $R$ remain in BF16.

Unlike standard GQA where keys and values are quantized separately along their respective reduction
axes, MLA reuses the single latent cache $C$ for both the score ($q_c^\top C^\top$) and context
($P C$) products. Because the microscaling groups of $\hat{C}$ are aligned along the channel
dimension $d_c$ for the FP4 score GEMM, executing the context reduction on FP4 Tensor Cores would
require re-quantizing $C$ along the token dimension at runtime. Instead, we dequantize the current
latent block $\tilde{C}_j = \phi^{-1}(s_{C_j}, \hat{C}_j)$ on the fly and accumulate the weighted
latent in FP16, omitting the probability quantization of Section~\ref{sec:p-scaling}. Finally,
because $(C - \mu_c)H^\top$ is used in the value path as well, we invert the Hadamard rotation and
restore the latent mean $\mu_c$ before applying the value up-projection $W_V$ (in practice, $H$ can
be folded offline into $W_K$ and $W_V$).

\end{document}